\documentclass[10pt,twocolumn,letterpaper]{article}

\usepackage{wacv}                       % CAMERA-READY version
\usepackage{times}
\usepackage{latexsym}
\usepackage[T1]{fontenc}
\usepackage[utf8]{inputenc}
\usepackage{microtype}
\usepackage{inconsolata}

\usepackage{graphicx}
\usepackage{booktabs}
\usepackage{amsfonts}
\usepackage{amsmath}
\usepackage{amssymb}
\usepackage{nicefrac}
\usepackage{xcolor}
\usepackage{tabularx}
\usepackage{multirow}
\usepackage{algorithm}
\usepackage{algorithmic}
\usepackage{url}

\graphicspath{{figures/}}

\definecolor{wacvblue}{rgb}{0.21,0.49,0.74}
\usepackage[pagebackref,breaklinks,colorlinks,allcolors=wacvblue]{hyperref}

\def\wacvPaperID{2535} % *** Enter the WACV Paper ID here
\def\confName{WACV}
\def\confYear{2027}

\title{T2I Models Rarely Invert Negation but Often Fail to Suppress Negated Concepts: Benchmarking Negation Suppression in Text-to-Image Generation}

\author{
Akansha Gautam$^{1}$,
Aparna Bharati$^{2}$,
Mayank Vatsa$^{1}$,
Richa Singh$^{1}$\\
$^{1}$Indian Institute of Technology Jodhpur, India \quad
$^{2}$Colorado State University, USA\\
{\small\ttfamily m23csa506@alumni.iitj.ac.in \quad
aparna.bharati@colostate.edu} \quad
{\small\ttfamily \{mvatsa,richa\}@iitj.ac.in}
}

\begin{document}
\maketitle

\begin{abstract}

Text-to-image diffusion models often fail to suppress explicitly negated concepts, such as a lamp for ``a desk without a lamp.'' We introduce \textbf{T2I-NSI}, a benchmark and metric for measuring negation sensitivity in text-to-image generation. It pairs each negated prompt with a canonical positive counterpart and factorizes negation by linguistic form (lexical/semantic, morphological, syntactic) and target type (objects, attributes, relations, compositions, and nested cases). The benchmark contains $759{,}869$ paired prompt atoms over a $5{,}082$-object catalog, with $916{,}272$ images ($911{,}982$ open-weights, $4{,}290$ closed-source), scored by a Basic CLIP estimator and a Robust ensemble of CLIP, VQA and open-vocabulary detection. Negation sensitivity is low: open-weights generators score NSI $\leq 0.038$, while GPT-Image-1.5 and Gemini-2.5-Flash-Image reach about $0.09$. The dominant failure is under-suppression rather than misdirection. Inverse interpretation is essentially absent ($0.00\%$ across over $450$K atom-model evaluations), whereas models render the concept under both prompts on $8$-$19\%$ of atoms and only partly suppress it on a further $9$-$25\%$; most remaining atoms are too weakly rendered to label. A blinded human study ($420$ atoms, nine annotators) corroborates the near-absence of inversion and the high persistence rate, and a matched-prompt comparison shows the open-versus-closed gap ($\sim\!3.5\times$) is not a prompt-generator artifact. \url{https://github.com/IAB-IITJ/T2I-NSI}
\end{abstract}

\begin{figure*}[t]
    \centering
    \includegraphics[width=1.0\textwidth]{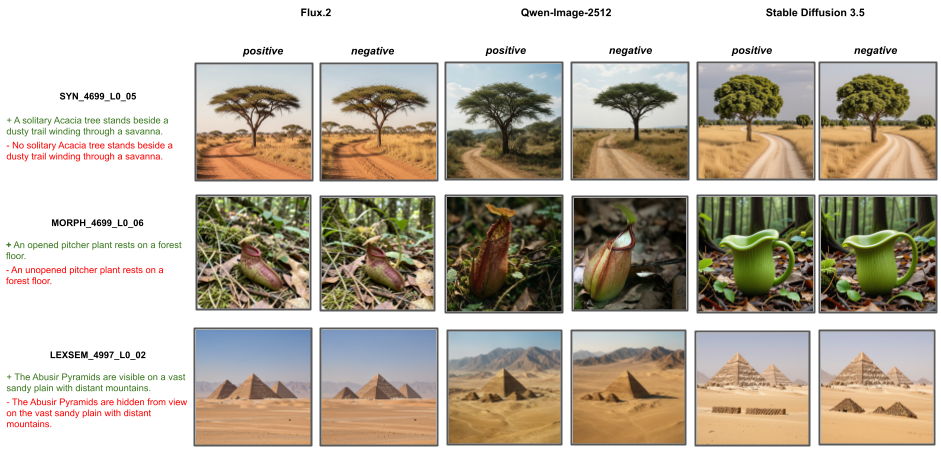}
\caption{\textbf{T2I generators fail to suppress negated concepts.} Three negation atoms spanning form: (top) syntactic (``No solitary Acacia tree\ldots''), (middle) morphological (``unopened pitcher plant''), and (bottom) lexical/semantic (``Abusir Pyramids hidden from view''). Across FLUX.2, Qwen-Image-2512, and SD~3.5, the negated prompt yields a near-identical image, and the negated concept (tree, opened pitcher, visible pyramids) persists.}
    \label{fig:teaser}
    \vspace{-15pt}
\end{figure*}

% =============================================================================
\section{Introduction}
\label{sec:intro}

Modern text-to-image (T2I) diffusion models render complex scenes with multiple objects, plausible spatial layouts, and tight attribute binding. Yet they routinely fail on inputs that should be linguistically simple: prompts like ``a desk \emph{without} a lamp'', ``an \emph{unlit} lamp on the table'', or ``\emph{no} streetlights in the village'' often produce images that still contain the negated object. This behavior persists across both open-weights and closed-source frontier generators, paralleling the negation blindness recently documented in vision-language models on the discriminative side~\cite{alhamoud2025vision,zhang2025negvqa,singh2025learning,park2025know,ko2025bringing}. The consequences are more severe in the generative setting: the user has no token-level output to inspect, only the wrong image. The risk is highest where a deployment must \emph{exclude} a concept, such as a child-free advertisement, a logo-free product shot or a weapon-free scene. Silently inserting the negated content there is a failure no scalar fidelity metric detects (Figure~\ref{fig:teaser}).

Negation is linguistically heterogeneous, so one aggregate score hides the failure modes that matter. Its \emph{form} may be \emph{lexical/semantic} (``dirty'' for ``clean''), \emph{morphological} (``\emph{un}lit'') or \emph{syntactic} (``\emph{no} streetlights'')~\cite{negation_taxonomy_2025}. Its \emph{target} ranges from a single object to attributes, relations and compositions, so a faithful evaluation must factorize across both axes. T2I suites such as T2I-CompBench~\cite{huang2023t2icompbench}, GenEval~\cite{ghosh2023geneval}, DALL-Eval~\cite{cho2023dalleval} and TIFA~\cite{hu2023tifa} cover attribute binding, spatial relations, counting and compositionality but omit negation. Negation benchmarks such as CC-Neg~\cite{singh2025learning}, NegBench~\cite{alhamoud2025vision}, NegVQA~\cite{zhang2025negvqa}, Thunder-NUBench~\cite{so2025thunder} and ``This is not a Dataset''~\cite{garcia-ferrero-etal-2023-dataset} instead target the discriminative side (retrieval, VQA, NLI). Recent work probes negation in T2I directly~\cite{conwell2024relations,hayes2025finegrain}, but with at most a few hundred hand-crafted prompts and without factorizing negation by form or target. \emph{To our knowledge, T2I-NSI is the largest benchmark for linguistic negation suppression in T2I generation, and the first to factorize it by form and target}. It combines benchmark-scale evaluation, systematic form $\times$ target factorization, automated probe scoring, failure-mode decomposition, and matched open/closed generator comparison.

Concretely, \textbf{T2I-NSI} comprises $759{,}869$ prompt atoms from a $5{,}082$-object catalog, factorized along form $\times$ target. Its black-box metric has two variants: \emph{Basic} (one CLIP probe) and \emph{Robust} (a CLIP/VQA/detection ensemble with uncertainty weighting, realization filtering, the preservation-aware \emph{NegAlign} and a per-atom failure mode). Every negated prompt is paired with a canonical positive counterpart, so suppression is measured by \emph{contrast} rather than by parsing the negated prompt. Probes are queried only with positive concept descriptions, reducing exposure to probe-side negation blindness~\cite{alhamoud2025vision,zhang2025negvqa,park2025know,singh2025learning}. This does not remove probe-side confounds, so a score is a probe-based measurement, not a direct estimate of \emph{generator} sensitivity. A \emph{negated prompt} here is an ordinary instruction whose text contains linguistic negation, distinct from the \emph{negative-prompt} conditioning field of many diffusion interfaces.

Across six generators, T2I-NSI surfaces an open-versus-closed gap that existing suites cannot see. Under the Robust estimator, the three open-weights generators score $\mathrm{NSI}\!\leq\!0.038$ and render the negated concept regardless of the prompt on $13$-$17\%$ of atoms (\texttt{ALWAYS\_PRESENT}). GPT-Image-1.5 and Gemini-2.5-Flash-Image reach $\mathrm{NSI}\!\approx\!0.09$, $3.3$-$3.5\!\times$ the open tier on identical prompts (Section~\ref{sec:experiments}). Yet across all six generators the \texttt{INVERSE} rate (target present only when negated) is essentially $0.00\%$: the dominant failure is \emph{under-suppression}, not inverse interpretation.

\noindent\textbf{Contributions:} We release
\begin{itemize}
  \item the $759{,}869$-atom paired-prompt corpus with canonical positive anchors;
  \item the Basic and Robust probe-based estimators, \emph{NegAlign} and a six-way failure-mode taxonomy;
  \item the largest evaluation of negation suppression across open- and closed-source generators to date, comprising $916{,}272$ main-benchmark images on six models; and
  \item the finding that \emph{under-suppression}, not misdirection, is the dominant failure mode, supported by human-validation, threshold, filter, seed-variance and matched-prompt analyses that also expose the metric's calibration limits.
\end{itemize}

%\noindent\textbf{Contributions:} We release \textbf{(i)} the $759{,}869$-atom paired-prompt corpus with canonical positive anchors; \textbf{(ii)} the Basic and Robust probe-based estimators, \emph{NegAlign} and a six-way failure-mode taxonomy; \textbf{(iii)} the first benchmark-scale evaluation across open- and closed-source generators, comprising $916{,}272$ images on six models; \textbf{(iv)} the finding that \emph{under-suppression}, not misdirection, is the dominant failure mode; and \textbf{(v)} human-validation, threshold, filter, seed-variance and matched-prompt analyses that test these findings and expose the metric's calibration limits.

%

% =============================================================================
\section{Related Work}
\label{sec:related_work}

\begin{table*}[t]
\centering
\footnotesize
\setlength{\tabcolsep}{4pt}
\caption{\textbf{Comparison with prior work} (NegBench~\citep{alhamoud2025vision}, Conwell et al.~\citep{conwell2024relations}, Sammani et al.~\citep{sammani2026geometry}, FineGRAIN~\citep{hayes2025finegrain}). T2I-NSI uniquely combines generation-side scope, factorized axes, a probe-ensemble metric, open/closed generators, and per-atom failure-mode classification.}
\label{tab:related_compare}
\vspace{-5pt}
\begin{tabular}{l c c c l c c}
\toprule
\textbf{Work} & \textbf{T2I} & \textbf{Scale} & \textbf{Factorized} & \textbf{Metric} & \textbf{Generators} & \textbf{Failure modes} \\
\midrule
NegBench           & $\times$    & $79$K                & partial               & retrieval acc.       & N/A          & $\times$ \\
Conwell~et~al.    & \checkmark  & few hundred          & $\times$              & human ratings        & open+closed  & $\times$ \\
Sammani~et~al.    & $\times$    & qualitative          & $\times$              & geometric            & N/A          & $\times$ \\
FineGRAIN          & \checkmark  & ${\sim}750$ prompts  & $\times$              & VLM judge            & open only    & 27-way \\
\textbf{T2I-NSI (ours)} & \checkmark  & $\mathbf{759{,}869}$ (152K eval.) & \checkmark\ (form$\times$target) & 3-probe + NegAlign & \checkmark\ (3+3) & 6-way \\
\bottomrule
\end{tabular}
\vspace{-10pt}
\end{table*}

\noindent\textbf{Negation in language and vision-language models:}
Large language and vision-language models systematically mishandle negation. \emph{This is not a Dataset}~\citep{garcia-ferrero-etal-2023-dataset} shows LLMs far below random on ${\sim}400$K sentences, \citet{so2025thunder} add sentence-level tests, and NegVQA~\citep{zhang2025negvqa} reports a U-shaped scaling trend over $7{,}379$ items. Multimodally, CC-Neg/CoN-CLIP~\citep{singh2025learning} probe retrieval on $228{,}246$ negated pairs, and NegBench~\citep{alhamoud2025vision} finds near-chance VLM performance on $79$K tasks. \citet{bhatia2025learning} study the related exclusion-retrieval setting, and the taxonomy of \citet{negation_taxonomy_2025} motivates our form $\times$ target factorization. Mitigations span data augmentation, training objectives, architectures, subspace analyses, and clinical adaptations~\citep{singh-etal-2023-nlms,park2025know,cai2025tngclip,rezaei2025making,xiao2026clipglasses,ranjbar2025spacevlm,ko2025bringing}. These efforts focus on the \emph{discriminative} side; generation is touched only through small-scale probes.

\noindent\textbf{Negation in text-to-image generation:}
Generation-side studies of negation remain limited. \citet{conwell2024relations} evaluate DALL-E~3 and grounded diffusion on relations, negations, and numbers, finding that negation prompts fail most often and no model exceeds $50\%$ human agreement. Their study, however, uses a small set of hand-crafted probes and no automated metric. \citet{sammani2026geometry} find a steerable negation direction in CLIP embedding space, but study retrieval rather than generation. FineGRAIN~\citep{hayes2025finegrain} scores $27$ failure modes with a VLM judge. Negation is one of them, but it is covered by only about $30$ hand-written prompts and is not factorized by linguistic form or target.

\noindent\textbf{T2I evaluation more broadly:}
Broad T2I suites (T2I-CompBench~\citep{huang2023t2icompbench}, GenEval~\citep{ghosh2023geneval}, DALL-Eval~\citep{cho2023dalleval}, TIFA~\citep{hu2023tifa}) measure compositionality, binding, counting and spatial relations via CLIPScore~\citep{hessel2021clipscore}, detection or VQA probes~\citep{singh2023divide}. None treats negation as a core axis, nor distinguishes whether a concept is always present or simply never rendered, a critical distinction our Robust estimator explicitly surfaces (Section~\ref{sec:experiments}).

\noindent\textbf{How T2I-NSI differs:}
Table~\ref{tab:related_compare} contrasts T2I-NSI with prior work across scale, factorization, model coverage and metric design. Two design choices set it apart: probes are queried only with positive concepts, and failures are scored with a preservation-aware alignment score and a six-way failure-mode taxonomy. Sections~\ref{sec:probes} and~\ref{sec:scoring} define them.

% =============================================================================

\section{Benchmark Construction}
\label{sec:benchmark}

T2I-NSI requires $\mathcal{O}(10^6)$ paired (positive, negative) prompt atoms so that the form $\times$ target factorization is well-populated for every model under test. Frontier closed-source LLMs produce the highest-quality negation prompts but were impractical for corpus generation at this scale because of cost and throughput constraints. Small open-weights models, in contrast, routinely violate the schema. For example, they fall back to syntactic ``not'' when a morphological prefix is requested. We therefore ran a controlled pilot that scored seven candidates on $7{,}500$ requested prompt pairs each. It selected DeepSeek~v3.2~UD-Q4\_K\_XL~\citep{deepseekv32} for the full open-weights corpus and Claude Sonnet~4.6~\citep{claudesonnet46} for the closed-source extension.

\noindent\textbf{Object catalog:}
The benchmark is seeded by a $5{,}082$-object catalog assembled from a manually curated noun list and the visually grounded subset of Synthetic Visual Genome objects~\cite{park2025synthetic}. 
We prune the pool with a cyclic triple-judge protocol in which GPT, Gemini, and Claude rotate between primary-selector and judge roles, averaging selection bias across three pre-training distributions. The resulting catalog covers everyday objects, architectural elements, vehicles, foliage, apparel, tools, and culturally diverse artifacts.

\noindent\textbf{Open-weights prompt corpus:}
We run DeepSeek~v3.2 over the catalog at the design ceiling of $5{,}082\!\times\!3\!\times\!5\!\times\!10 = 762{,}300$ atoms (objects $\times$ negation classes $\times$ levels $\times$ samples). Schema validation leaves $\mathbf{759{,}869}$ valid atom rows. Of the $2{,}431$ missing rows, the largest share is morphological ($1{,}056$, vs.\ $851$ syntactic and $524$ lexical/semantic)\footnote{Morphological negation is a bound affix (\emph{un-}, \emph{-less}), so a model can satisfy the surface request while normalizing the affix back to the unnegated stem (``unlit''$\rightarrow$``lit''). All seven LLMs in the pilot also scored lowest on this form (supplementary, \emph{Prompt-Generator Selection Pilot}).}. After lowercasing and removing punctuation, $756{,}571$ prompt pairs ($99.6\%$) are unique. We release the full corpus and render images on the subset described next.

\noindent\textbf{Open-weights image-rendering subset:}
Image generation runs on a balanced subset that gives every form $\times$ target cell enough coverage for stable conditioned-score estimation. Sampling about $5{,}000$ objects across $3$ classes, $5$ levels, and $2$ samples yields $\mathbf{151{,}997}$ atoms, about $10{,}000$ per cell. With one positive and one negated prompt per atom, this gives $\mathbf{303{,}994}$ rendered prompts. We use $K{=}1$ image per prompt with paired seeds, so the positive and negated prompts of an atom share one seed. Rendering across FLUX.2, Qwen-Image-2512, and SD~3.5 yields $\mathbf{911{,}982}$ images.

\noindent\textbf{Closed-source extension:}
To test whether closed-source systems share this insensitivity, we build a matched stratified sample for GPT-Image-1.5, Gemini-2.5-Flash-Image and Midjourney~v7, sized to cover every form $\times$ level cell rather than catalog breadth: $\mathbf{1{,}200}$ atoms ($80$ per cell) and $\mathbf{2{,}400}$ prompts, generated by Claude Sonnet~4.6. API rate limits and cost restrict coverage. GPT-Image-1.5 renders all $1{,}200$ atoms, Gemini-2.5-Flash-Image $717$, and Midjourney~v7 a $228$-atom syntactic-only slice, for $\mathbf{4{,}290}$ closed-source images. Midjourney~v7 exposes no public API, so we submitted each pair to the Midjourney bot in a private Discord channel and scored the first quadrant of each $2\!\times\!2$ grid, with one shared \texttt{-{}-seed} per atom. Because these prompts come from a different LLM, we compare the tiers directly only on identical prompts (Section~\ref{sec:experiments}).

\noindent\textbf{Dataset structure and atom schema:}
Each atom is one row of the prompt files, recording provenance (\texttt{atom\_id}, seed \texttt{object}), the factorization axes (\texttt{negation\_class}, \texttt{complexity\_level}, \texttt{target\_type}) and the prompt pair. Per-generator image-metadata files add, for rendered atoms, the target sets $T_j^{\text{neg}}$/$T_j^{\text{pres}}$, image paths and seeds. A second DeepSeek pass fills the target sets with canonical positive noun phrases and drops non-visual predicates via a visuality flag. The supplementary gives the full schema and worked examples (\emph{Atom Schema and Worked Examples}) and the Atoms $\to$ Prompts $\to$ Images accounting (\emph{Generation Provenance}).

% =============================================================================
\section{T2I-NSI: Metric for Negation}
\label{sec:metric}

\begin{algorithm*}[t]
\caption{Robust T2I-NSI: per-atom scoring and global aggregation.}
\label{alg:robust_nsi}
\small
\begin{algorithmic}[1]
\REQUIRE Atoms $\mathcal{A} = \{a_j\}_{j=1}^J$, model $\mathcal{M}$, probes $\{g_e\}_{e=1}^E$, image encoder $\phi$, samples $K$; thresholds $\xi, \theta_{\text{pos}}, \theta_{\text{neg}}$; scale $\beta$.
\FOR{each atom $a_j \in \mathcal{A}$}
    \STATE Sample $p_j^+ \!\in\! \mathcal{P}_j^+$, $p_j^- \!\in\! \mathcal{P}_j^-$ and seeds $\{z_k\}_{k=1}^K$; render $I_{j,k}^\pm \!=\! \mathcal{M}(p_j^\pm, z_k)$ \COMMENT{paired $z_k$ where seed control exists; else independent samples}
    \STATE \textbf{Target scoring.} For $t \!\in\! T_j^{\text{neg}}, k$: compute $\bar{s}_{k,t}^\pm \!=\! \mathrm{median}_e g_e(I_{j,k}^\pm, t)$; let $u_k \!=\! \max_{t,\pm} \mathrm{Var}_e\, g_e(I_{j,k}^\pm, t)$ be the largest probe disagreement at sample $k$.
    \STATE Aggregate: $R_j^\pm \!\leftarrow\! \mathrm{mean}_{k,t}\bar{s}_{k,t}^\pm$;\, $C_j \!\leftarrow\! R_j^+$;\, $U_j \!\leftarrow\! \mathrm{mean}_{k} u_k$;\, $w_j \!\leftarrow\! \exp(-\beta U_j)$.
    \STATE \textbf{Gated preservation.} For $t \!\in\! T_j^{\text{pres}}, k$ with $\mathrm{median}_e g_e(I_{j,k}^+, t) \!\geq\! \theta_{\text{pos}}$: accumulate $\mathrm{median}_e g_e(I_{j,k}^-, t)$ into $P_j^-$; set $P_j^- \!\leftarrow\! 1$ if no $(k, t)$ qualifies.
    \STATE \textbf{Scores.} $\mathrm{NSI}_j \!\leftarrow\! R_j^+ \!-\! R_j^-$;\, $\mathrm{NegAlign}_j \!\leftarrow\! \tfrac{1}{2}(C_j + (1 \!-\! R_j^-))\, P_j^-$;\, $S_j \!\leftarrow\! \mathrm{mean}_k \mathrm{sim}_{\text{img}}(I_{j,k}^+, I_{j,k}^-)$.
    \STATE \textbf{Failure mode.} Assign $\mathrm{class}_j \!\in\! \{\texttt{SUCCESS}, \texttt{ALWAYS\_PRESENT}, \texttt{ALWAYS\_ABSENT}, \texttt{INVERSE}, \texttt{PARTIAL\_SUPPR}, \texttt{WEAKLY\_REND}\}$ from $(C_j, R_j^-)$ via \S\ref{sec:scoring}.
\ENDFOR
\STATE Filter $\mathcal{J} \!\leftarrow\! \{j \!:\! C_j \!\geq\! \xi\}$;\, $\mathrm{NSI}(\mathcal{M}) \!\leftarrow\! \sum_{j \in \mathcal{J}} w_j \mathrm{NSI}_j \big/ \sum_{j \in \mathcal{J}} w_j$.
\STATE Compute conditioned scores $\mathrm{NSI}_{\tau^{\text{target}}}, \mathrm{NSI}_{\tau^{\text{form}}}, \mathrm{NSI}_{\tau^{\text{target}}, \tau^{\text{form}}}$ over the corresponding atom subsets.
\RETURN $\mathrm{NSI}(\mathcal{M})$, conditioned scores, per-atom $\{(\mathrm{NSI}_j, C_j, \mathrm{NegAlign}_j, w_j, \mathrm{class}_j, S_j, U_j)\}_{j}$.
\end{algorithmic}
\end{algorithm*}

T2I-NSI quantifies a model's ability to respect negated instructions by treating negation as a contrastive switch. Ideally, a target concept should be realized under positive prompts and suppressed under negation, while the rest of the scene remains coherent. The metric is black-box: it evaluates generated images through external vision-language probes without access to internal model weights.

\subsection{Atoms, Probes, and Probe-Side Isolation}
\label{sec:probes}

Let $\mathcal{M}$ be a T2I model that maps a prompt $p$ and seed $z$ to an image $I = \mathcal{M}(p, z)$. The benchmark is a set of negation atoms $\mathcal{A} = \{a_j\}_{j=1}^J$. Each atom supplies positive and negated prompt sets $\mathcal{P}_j^{\pm}$ along with a set $T_j$ of visual concepts and a non-empty negation target set $T_j^{\text{neg}} \subseteq T_j$. It provides a preservation set $T_j^{\text{pres}} = T_j \setminus T_j^{\text{neg}}$ of visual concepts that must remain present under negation. We write $n_j = |T_j^{\text{neg}}|$ and use $t_j^{*}$ for the unique target when $n_j = 1$. A factorization tag $(\tau_j^{\text{target}}, \tau_j^{\text{form}})$ records the dominant target and linguistic form.

\noindent\textbf{Probe ensemble:} T2I-NSI assumes access to $E \!\geq\! 1$ vision-language probes $g_e: \mathcal{I} \times \mathcal{T} \to [0,1]$, where $g_e(I, t)$ scores the confidence that concept $t$ holds in image $I$, aggregated by median $g(I, t) := \mathrm{median}_e\, g_e(I, t)$. The Basic estimator uses $E\!=\!1$ with CLIP-L/14~\citep{radford2021clip}. The Robust estimator scales to an ensemble of $E\!=\!3$ probes: CLIP-L/14, InstructBLIP~\citep{dai2023instructblip}, and GroundingDINO~\citep{liu2024grounding}. These are complementary along global-semantic, fine-grained-VQA, and localized-detection axes. A CLIP image encoder $\phi$ defines paired-image similarity $\mathrm{sim}_{\text{img}}(I, I') = \langle \phi(I), \phi(I') \rangle / (\|\phi(I)\|\,\|\phi(I')\|)$.

\noindent\textbf{Probe-side isolation:} A core design principle is that probes are queried using only \emph{positive} concept descriptions, regardless of how the prompt expresses negation. For example, ``lit lamp'' anchors ``unlit lamp,'' and ``all flowers are vibrant'' is the canonical paraphrase for ``no flowers are not vibrant.'' This reduces the metric's exposure to the probe-side negation blindness documented in the VLM literature~\citep{alhamoud2025vision,zhang2025negvqa,park2025know,singh2025learning}. Lexical and morphological atoms use a canonical positive anchor for each negated concept, and stacked syntactic forms use logical paraphrases; each probe receives the same positive query for both images of a pair. Linguistic complexity therefore lives in the generation prompt rather than the probe query. Full normalization rules are in the supplementary (\emph{Metric Details and Algorithms}).

\subsection{Scoring: Sensitivity, Alignment, and Failure Modes}
\label{sec:scoring}

\noindent\textbf{Negation Sensitivity Index:} For each atom $a_j$, we sample $p_j^+ \in \mathcal{P}_j^+$ and $p_j^- \in \mathcal{P}_j^-$, and render images $I_{j,k}^{\pm} = \mathcal{M}(p_j^{\pm}, z_k)$ using paired seeds $\{z_k\}_{k=1}^K$ \emph{where the generator exposes seed control}. For providers that do not, $I_{j,k}^{+}$ and $I_{j,k}^{-}$ are independent samples from the matched prompt pair and $z_k$ is undefined. For each target $t \in T_j^{\text{neg}}$, we estimate per-target presence $R_{j,t}^{\pm} = \tfrac{1}{K}\sum_{k} g(I_{j,k}^{\pm}, t)$ and per-target sensitivity $\mathrm{NSI}_{j,t} = R_{j,t}^{+} - R_{j,t}^{-}$. Atom-level scores average across targets, $R_j^{\pm} = \tfrac{1}{n_j}\sum_{t \in T_j^{\text{neg}}} R_{j,t}^{\pm}$, yielding the sensitivity index:
\begin{equation}
\mathrm{NSI}_j = R_j^{+} - R_j^{-} \in [-1, 1].
\end{equation}
Values close to $1$ indicate that the target appears under positive prompts and is suppressed under negation, while values near $0$ or below require the diagnostics defined next. The aggregate $\mathrm{NSI}_j$ alone cannot separate a model that never generates the concept from one that generates it but ignores the negation, so low NSI does not by itself show that a model misunderstood negation. We therefore compute three diagnostics that decompose the score.

\noindent\textbf{Positive realization and filter retention:} The value $C_j = R_j^{+}$ is the \emph{positive realization} of the negation targets: how strongly the model renders them when asked to. Atoms with $C_j < \xi$ for a realization threshold $\xi$ cannot reliably support NSI inference, so we report realization-filtered scores $\mathrm{NSI}_{C \geq \xi}$ restricted to atoms above the threshold, and call the surviving fraction the \emph{filter retention}. The two are distinct. Mean $C_j\!\approx\!0.64$ is an average \emph{strength} of realization, whereas filter retention $\approx\!99.4\%$ is the fraction of atoms clearing the much lower bar $C_j\!\geq\!\xi$. Most atoms are realized well enough to be scored, even though few are realized strongly.

\noindent\textbf{Preservation and negation alignment:} Preservation $P_j^{-} = \frac{1}{|T_j^{\text{pres}}|}\sum_{t \in T_j^{\text{pres}}} \frac{1}{K}\sum_{k} g(I_{j,k}^{-}, t)$ measures whether the background context $T_j^{\text{pres}}$ survives the negated prompt ($P_j^{-}\!=\!1$ when $|T_j^{\text{pres}}|\!=\!0$). In the Robust estimator it uses only $(k,t)$ pairs confidently realized in the positive image ($\mathrm{median}_e\, g_e(I_{j,k}^{+}, t) \geq \theta_{\text{pos}}$), separating preservation failures from realization failures. The three signals combine into one preservation-aware score,
\begin{equation}
    \mathrm{NegAlign}_j = \tfrac{1}{2}\big(C_j + (1 - R_j^{-})\big)\, P_j^{-},
\end{equation}
which is high when the model realizes targets positively, suppresses them under negation, and preserves the surrounding scene.

\noindent\textbf{Failure-mode taxonomy:} Because the scalar $\mathrm{NSI}_j$ collapses qualitatively distinct behaviors, we also label each atom from $(C_j, R_j^{-})$ using two thresholds $\theta_{\text{neg}} \!<\! \theta_{\text{pos}}$. A score is \emph{high} if it is $\geq\! \theta_{\text{pos}}$ and \emph{low} if it is $\leq\! \theta_{\text{neg}}$. \texttt{SUCCESS} (high $C_j$, low $R_j^{-}$) denotes targets rendered under positive prompts but suppressed under negation. \texttt{ALWAYS\_PRESENT} (high $C_j$, high $R_j^{-}$) captures insensitivity, where the concept appears regardless of the prompt. \texttt{ALWAYS\_ABSENT} (low $C_j$, low $R_j^{-}$) marks a capability gap, where the model rarely renders the concept in either condition. \texttt{INVERSE} (low $C_j$, high $R_j^{-}$) marks misdirection, where the concept appears only under negation. All remaining atoms form a fifth, residual \texttt{OTHER} region, reported as two subclasses. \texttt{PARTIAL\_SUPPR} (high $C_j$, $\theta_{\text{neg}} \!<\! R_j^{-} \!<\! \theta_{\text{pos}}$) means the target is realized but only partly attenuated. \texttt{WEAKLY\_REND} ($C_j \!<\! \theta_{\text{pos}}$, no other label) means the realization score is below threshold. The six labels partition the atoms. These labels are thresholded probe scores, not human-equivalent categories. Per-class counts are key to our findings (Section~\ref{sec:experiments}).

\noindent\textbf{Robust estimator:} Algorithm~\ref{alg:robust_nsi} folds probe ensembling, uncertainty bookkeeping, gated preservation, and failure-mode classification into a single per-atom pass, followed by realization-filtered weighted aggregation with $w_j \!=\! \exp(-\beta U_j)$ for a scale $\beta\!>\!0$. The Basic estimator drops these extensions. Full pseudocode, complexity derivations, and normalization rules are in the supplementary (\emph{Metric Details and Algorithms}).

% =============================================================================

\section{Experimental Analysis}
\label{sec:experiments}

We evaluate T2I-NSI on six text-to-image generators, three open-weights and three closed-source. We report the open-weights tier at the full $\sim\!152$K-atom scale and the closed-source extension on per-API matched subsets. The supplementary reports the \emph{Basic vs.\ Robust Agreement} analysis and the \emph{Closed-Source Conditioned Profile}.

\begin{table*}[t]
\centering
\small
\setlength{\tabcolsep}{4.5pt}
\caption{\textbf{Aggregate T2I-NSI on the open-weights tier} ($\sim\!152$K atoms per model). The metric is bounded in $[-1, 1]$; a perfectly negation-aware generator scores $\mathrm{NSI}\!=\!1$. ``Mean $C_j$'' is the mean positive realization $C_j\!=\!R_j^+$ (not a retention rate). The Robust estimator additionally reports the mean ensemble weight $w_j\!=\!\exp(-\beta U_j)$, the global paired-image similarity $S_j$, and mean per-atom uncertainty $U_j$; these columns are not produced by the Basic estimator.}
\label{tab:headline}
\begin{tabular}{l l c c c c c c}
\toprule
\textbf{Algorithm} & \textbf{Model} & $\mathrm{NSI}(\mathcal{M})$ & Mean $C_j$ & NegAlign & Weight & Similarity & Uncertainty \\
\midrule
\multirow{3}{*}{Basic (CLIP)}
  & FLUX.2          & $0.0035$ & $0.6095$ & $0.3007$ & --- & --- & --- \\
  & Qwen-Image-2512 & $0.0036$ & $0.6115$ & $0.3009$ & --- & --- & --- \\
  & SD~3.5          & $0.0017$ & $0.6118$ & $0.3000$ & --- & --- & --- \\
\midrule
\multirow{3}{*}{Robust (3-probe)}
  & FLUX.2          & $\mathbf{0.0378}$ & $0.6477$ & $\mathbf{0.4617}$ & $0.9367$ & $0.9557$ & $0.0333$ \\
  & Qwen-Image-2512 & $0.0306$          & $0.6404$ & $0.4612$          & $0.9297$ & $0.9343$ & $0.0372$ \\
  & SD~3.5          & $0.0198$          & $0.6402$ & $0.4548$          & $0.9413$ & $0.9633$ & $0.0308$ \\
\bottomrule
\end{tabular}
\end{table*}

\begin{table*}[t]
\centering
\small
\setlength{\tabcolsep}{6pt}
\caption{\textbf{Per-atom failure-mode distribution} for the open-weights generators under the Robust estimator, with classes as defined in Section~\ref{sec:scoring}. Rows sum to $100\%$ (rounding aside); \texttt{INVERSE} is $0.00\%$ to 2\,dp ($0$/$3$/$2$ atoms of $151{,}970$).}
\label{tab:failure}
\begin{tabular}{l c c c c c c}
\toprule
\textbf{Model} & \texttt{SUCCESS} & \texttt{ALWAYS\_PRESENT} & \texttt{ALWAYS\_ABSENT} & \texttt{INVERSE} & \texttt{PARTIAL\_SUPPR} & \texttt{WEAKLY\_REND} \\
\midrule
FLUX.2           & $0.29\%$ & $15.50\%$ & $0.31\%$ & $0.00\%$ & $\mathbf{12.09\%}$ & $71.81\%$ \\
Qwen-Image-2512  & $0.06\%$ & $13.03\%$ & $0.19\%$ & $0.00\%$ & $12.54\%$ & $\mathbf{74.18\%}$ \\
SD~3.5           & $0.03\%$ & $\mathbf{16.92\%}$ & $0.31\%$ & $0.00\%$ & $9.16\%$ & $73.59\%$ \\
\bottomrule
\end{tabular}
\vspace{-10pt}
\end{table*}

\begin{figure}[!t]
\centering
\includegraphics[width=1.0\linewidth]{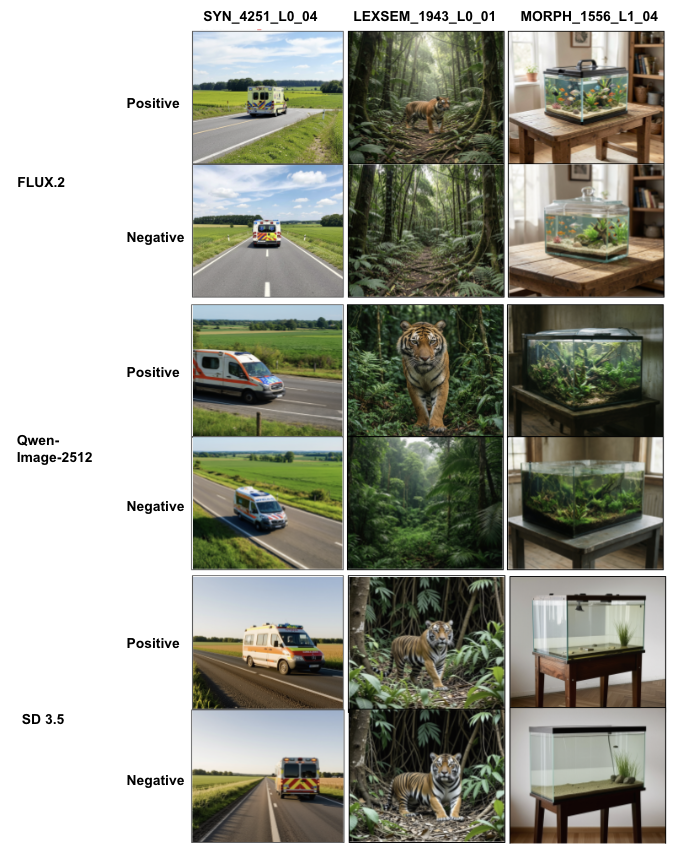}
\caption{\textbf{Representative paired generations: suppression varies by atom and is often incomplete.}
Positive (upper) and negated (lower) renders under matched seeds for FLUX.2, Qwen-Image-2512 and SD~3.5; column headers give the atom ids.
Prompts, positive~$\rightarrow$~negated. \emph{Left} (syntactic, L0): ``An ambulance is on the road on a rural highway with green fields.''~$\rightarrow$~``There is \emph{not} an ambulance on the road\ldots''
\emph{Middle} (lexical/semantic, L0): ``A tiger is \emph{present} in a dense jungle clearing.''~$\rightarrow$~``The tiger is \emph{absent} from the dense jungle clearing.''
\emph{Right} (morphological, L1): ``A \emph{covered} fish tank rests on a sturdy table.''~$\rightarrow$~``An \emph{uncovered} fish tank\ldots''
Robust-estimator classes (FLUX.2\,/\,Qwen-Image-2512\,/\,SD~3.5). \emph{Left}: all \texttt{ALWAYS\_PRESENT}. \emph{Middle}: \texttt{SUCCESS}\,/\,\texttt{SUCCESS}\,/\,\texttt{ALWAYS\_PRESENT}. \emph{Right}: all \texttt{WEAKLY\_REND} ($C_j\!=\!0.63$/$0.65$/$0.62\!<\!\theta_{\text{pos}}$).}
\label{fig:qualitative}
\vspace{-12pt}
\end{figure}

\noindent\textbf{Models, Probes, and Scale:} The open-weights checkpoints are \textbf{FLUX.2} (klein-9B)~\citep{flux2024}, \textbf{Qwen-Image-2512}~\citep{qwenimage2025} and \textbf{Stable Diffusion~3.5 Large Turbo} (SD~3.5)~\citep{sd35turbo,esser2024scaling,rombach2022high}. The closed-source generators are \textbf{GPT-Image-1.5}~\citep{gptimage15}, \textbf{Gemini-2.5-Flash-Image}~\citep{geminiflashimage} and \textbf{Midjourney~v7}~\citep{midjourneyv7}. Of the $151{,}997$ open-weights atoms, $151{,}970$ are scored; $27$ are dropped because a probe returned no valid score. The Basic estimator uses CLIP-L/14~\citep{radford2021clip}, and the Robust estimator adds InstructBLIP~\citep{dai2023instructblip} and GroundingDINO~\citep{liu2024grounding}. We set $\xi\!=\!0.3$, $\theta_{\text{pos}}\!=\!0.7$, $\theta_{\text{neg}}\!=\!0.3$, $\beta\!=\!2.0$ and $K\!=\!1$. Seeds are paired where the generator exposes seed control (the open-weights models and Midjourney~v7); GPT-Image-1.5 and Gemini-2.5-Flash-Image are sampled independently from matched prompts. Exact checkpoints, settings and per-model counts are in the supplementary (\emph{Generation Provenance}).

% =============================================================================
%\subsection{Aggregate T2I-NSI: Open-Weights}
%\label{sec:exp_headline}

\noindent\textbf{Aggregate T2I-NSI (Open-Weights):}
Table~\ref{tab:headline} reports $\mathrm{NSI}(\mathcal{M})$ and diagnostics under both estimators. First, Basic NSI is at most $0.0036$ for all three models: CLIP similarity to the anchor saturates, so a single probe cannot detect suppression. The Robust ensemble gives $\text{FLUX.2}\,(0.0378)\!>\!\text{Qwen-Image-2512}\,(0.0306)\!>\!\text{SD~3.5}\,(0.0198)$, and $56.5$-$61.3\%$ of atoms score higher than under Basic (supplementary, \emph{Basic vs.\ Robust Agreement})\footnote{Near-neutral \texttt{WEAKLY\_REND} atoms pull the aggregate toward zero, so it should be read with the class shares, which give no single ordering: \texttt{SUCCESS} follows the aggregate, but Qwen-Image-2512 has the lowest \texttt{ALWAYS\_PRESENT} share (Table~\ref{tab:failure}).}. Second, paired image similarity $S_j\!=\!0.93$-$0.96$ shows that scene structure survives the negation flip, so \emph{the negated concept itself is rarely removed}. Third, mean weight $w_j\!\approx\!0.93$-$0.94$ reflects low cross-probe variance, not equal probe contribution: VQA carries most of the suppression signal (Limitations).

% =============================================================================
%\subsection{Failure-Mode Distribution: Open-Weights}
%\label{sec:exp_failures}

\noindent\textbf{Failure-Mode Distribution (Open-Weights):}
Table~\ref{tab:failure} gives the per-atom failure-mode distribution, with examples in Figure~\ref{fig:qualitative}. The finding is \emph{structural} and has four parts. \emph{(i) Models almost never invert negation:} \texttt{INVERSE} is $0.00\%$ for all three open-weights generators ($0$/$3$/$2$ of $\sim\!456$K atom-model evaluations). \emph{(ii) Clean successes are rare:} $0.03$-$0.29\%$ of atoms (FLUX.2 the most, $N\!=\!440$). \emph{(iii) Under-suppression dominates:} the negated concept scores high under both prompts in $13.0$-$16.9\%$ of atoms (\texttt{ALWAYS\_PRESENT}) and is attenuated but not removed in a further $9.2$-$12.5\%$ (\texttt{PARTIAL\_SUPPR}, mean $\mathrm{NSI}_j\!=\!0.17$-$0.19$). \emph{(iv) A large residual receives no success/failure label:} for $71.8$-$74.2\%$ of atoms the realization score falls below $\theta_{\text{pos}}$. This is partly a probe-calibration effect, since humans label many of the metric's \texttt{OTHER} atoms \texttt{ALWAYS\_PRESENT} (Human Validation). These atoms are near-neutral (mean $\mathrm{NSI}_j\!=\!0.004$-$0.017$) and, because $99.6\%$ of these \texttt{WEAKLY\_REND} atoms still clear the lower $C_j\!\geq\!\xi$ bar, they dominate the aggregate and hold it near zero. SD~3.5 has the largest \texttt{ALWAYS\_PRESENT} share ($16.92\%$) and the smallest \texttt{SUCCESS} share, consistent with its lowest Robust NSI.

% =============================================================================
%\subsection{Conditioned Profiles: Open-Weights}
%\label{sec:exp_profiles}

\begin{table*}[t]
\centering
\small
\setlength{\tabcolsep}{4pt}
\caption{\textbf{Conditioned T2I-NSI scores} for the open-weights generators under the Robust estimator. The benchmark is factorized by linguistic form and \emph{target complexity}: levels L0-L4 correspond to increasingly structured target types (object, attribute, relation, composition, nested), so the two are a single ordered axis rather than independent factors. Bold marks the per-row maximum within each axis.}
\vspace{-5pt}
\label{tab:conditioned}
\begin{tabular}{l c c c c c c c c}
\toprule
\textbf{Model} & \multicolumn{3}{c}{\textbf{Form}} & \multicolumn{5}{c}{\textbf{Complexity level}}\\
\cmidrule(lr){2-4}\cmidrule(lr){5-9}
 & lex. & morph. & syn. & L0 (obj) & L1 (attr) & L2 (rel) & L3 (comp) & L4 (nest) \\
\midrule
FLUX.2           & $\mathbf{0.0580}$ & $0.0291$ & $0.0267$ & $\mathbf{0.0511}$ & $0.0434$ & $0.0218$ & $0.0386$ & $0.0344$ \\
Qwen-Image-2512  & $\mathbf{0.0511}$ & $0.0302$ & $0.0110$ & $0.0185$ & $0.0399$ & $0.0170$ & $\mathbf{0.0407}$ & $0.0372$ \\
SD~3.5           & $\mathbf{0.0301}$ & $0.0172$ & $0.0120$ & $0.0173$ & $\mathbf{0.0292}$ & $0.0104$ & $0.0245$ & $0.0173$ \\
\bottomrule
\end{tabular}
\vspace{-5pt}
\end{table*}

\begin{table*}[t]
\centering
\small
\setlength{\tabcolsep}{3.5pt}
\caption{\textbf{Closed-source extension under the Robust estimator.} Sample sizes vary by API constraint; Midjourney~v7 is syntactic-only, so its $0.0267$ should be read against the open-weights \emph{syntactic} scores ($0.0110$-$0.0267$, Table~\ref{tab:conditioned}), not the all-form aggregates. Basic CLIP NSI saturates at $\le\!0.008$ across all three, and \texttt{INVERSE} and \texttt{ALWAYS\_ABSENT} remain $0.00\%$. Class shares sum to $100\%$ per row (rounding aside).}
\vspace{-5pt}
\label{tab:closed}
\begin{tabular}{l c c c c c c c c c}
\toprule
\textbf{Model} & $N$ & $\mathrm{NSI}(\mathcal{M})$ & NegAlign & \texttt{SUCCESS} & \shortstack{\texttt{ALWAYS\_}\\\texttt{PRESENT}} & \shortstack{\texttt{ALWAYS\_}\\\texttt{ABSENT}} & \texttt{INVERSE} & \shortstack{\texttt{PARTIAL\_}\\\texttt{SUPPR}} & \shortstack{\texttt{WEAKLY\_}\\\texttt{REND}} \\
\midrule
GPT-Image-1.5            & $1{,}200$ & $\mathbf{0.0934}$ & $\mathbf{0.479}$ & $1.58\%$ & $11.42\%$ & $0.00\%$ & $0.00\%$ & $\mathbf{24.83\%}$ & $62.17\%$ \\
Gemini-2.5-Flash-Image   & $717$     & $0.0902$          & $0.476$          & $\mathbf{1.95\%}$ & $7.67\%$ & $0.00\%$ & $0.00\%$ & $18.97\%$ & $\mathbf{71.41\%}$ \\
Midjourney~v7 (syn.\ only) & $228$    & $0.0267$          & $0.448$          & $0.00\%$ & $\mathbf{19.30\%}$ & $0.00\%$ & $0.00\%$ & $13.16\%$ & $67.54\%$ \\
\bottomrule
\end{tabular}
\vspace{-10pt}
\end{table*}

\noindent\textbf{Conditioned Profiles (Open-Weights):}
Table~\ref{tab:conditioned} reports conditioned T2I-NSI by linguistic form and complexity level L0-L4 ($C_j\!\geq\!\xi$). Two patterns hold for all three open-weights generators. First, \emph{negation sensitivity follows lexical/semantic $>$ morphological $>$ syntactic}. Lexical/semantic negation (``dirty'' for ``clean'') is easiest, since the negated prompt names a familiar antonym. Syntactic negation (``no streetlights'') is hardest, dropping to $0.0110$ for Qwen-Image-2512, which mirrors discriminative VLM results~\citep{alhamoud2025vision,zhang2025negvqa}. Second, \emph{relational negation is the consistent trough}: L2 (``not on the table'') is the minimum for all three, below half the L0 score for FLUX.2. L3 and L4 recover partially because they embed a clear single-concept target, giving the generation-side analogue of the U-shape NegVQA~\citep{zhang2025negvqa} reports. These patterns may also reflect probe sensitivity, because the probes differ in how reliably they recognize objects and relations.

% =============================================================================

%\subsection{Closed-Source Extension}
%\label{sec:exp_closed_source}

\noindent\textbf{Closed-Source Extension:}
Table~\ref{tab:closed} reports the closed-source generators on per-API subsets ($N\!=\!1{,}200, 717, 228$). Their prompts differ from the open tier's, so this comparison is indicative and is tested on identical prompts below. GPT-Image-1.5 and Gemini-2.5-Flash-Image reach $\mathrm{NSI}\!\approx\!0.09$, $2.5\!\times$ the open-weights ceiling of $0.038$, with $5$-$7\times$ the \texttt{SUCCESS} share of the best open model and more partial suppression ($19$-$25\%$ vs.\ $9$-$13\%$). Under-suppression still dominates: \texttt{ALWAYS\_PRESENT} plus \texttt{PARTIAL\_SUPPR} covers $27$-$36\%$ of their atoms, and \texttt{INVERSE} is $0.00\%$ across all six generators. \emph{Midjourney~v7 behaves like an open-weights generator}: its syntactic NSI ($0.0267$) sits atop the open-weights syntactic band ($0.0110$-$0.0267$), with the highest \texttt{ALWAYS\_PRESENT} share ($19.30\%$) and zero \texttt{SUCCESS}.

\noindent\textbf{Matched-Prompt Open-vs-Closed:} To remove the prompt-source confound (DeepSeek open, Claude closed), we re-rendered the $1{,}200$ closed-tier atoms on all three open models ($7{,}200$ images) and compared shared subsets with paired atom-bootstrap CIs ($B\!=\!5000$; supplementary, \emph{Matched-Prompt Open-vs-Closed Comparison}). On identical prompts GPT-Image-1.5 scores $0.093$ against an open-tier mean of $0.027$ (FLUX.2 $0.039$, Qwen-Image-2512 $0.024$, SD~3.5 $0.017$), a $3.5\!\times$ gap ($\Delta\!=\!-0.067$, $95\%$ CI $[-0.075,-0.059]$). Gemini-2.5-Flash-Image shows $3.3\!\times$ on its $717$ atoms, and the open tier exceeds neither model in any resample, so removing the confound \emph{widens} the gap. Midjourney~v7 is indistinguishable from the open tier ($\Delta\!=\!-0.004$, $[-0.016,+0.010]$). This remains a probe-based result, and GPT-Image-1.5 and Gemini-2.5-Flash-Image use unpaired sampling.

\noindent\textbf{Human Validation:} The study covers \textbf{420 atoms}: $300$ representative atoms judged on both FLUX.2 and Qwen-Image-2512 ($600$ atom-model pairs) and $120$ disjoint rare-class diagnostic atoms ($120$ pairs). Every target is judged in both images of a pair, giving $2{,}522$ image-concept trials, each rated by $3$ of the $9$ annotators ($7{,}466$ non-missing ratings). Annotators saw only the image and a canonical positive phrase. Human judgments \emph{strongly corroborate two findings and qualify two others}. \emph{(a) Near-absence of inversion:} \texttt{INVERSE} is $0.5\%$ (human; $3/600$, Wilson $95\%$ CI $[0.2,1.5]\%$) vs.\ $0.0\%$ (metric). \emph{(b) High persistence under negation:} the target remains visible in $87.3\%$ (human) vs.\ $85.3\%$ (metric) of negated images, and humans read \emph{more} residual presence ($R_j^-\!=\!0.78$ vs.\ $0.62$). \emph{(c) The absolute \texttt{SUCCESS} rate is conservative:} humans place $R_j^-$ below $0.3$ on $8.2\%$ of representative atoms against the metric's $1.0\%$ ($6.2\%$ vs.\ $0.5\%$ for the full \texttt{SUCCESS} label). A relaxed cutoff chosen after inspection ($R_j^-\!<\!0.5$) brings the first pair close ($12.7\%$ human vs.\ $14.7\%$ metric), suggesting a calibration offset, so the automated \texttt{SUCCESS} figure is not an estimate of human-perceived success. \emph{(d) Labels are not human-equivalent:} five-way agreement (\texttt{OTHER} subclasses merged) is $42\%$ (Cohen $\kappa\!=\!0.09$), and humans call $283$ of the metric's $502$ \texttt{OTHER} atoms \texttt{ALWAYS\_PRESENT} ($59.8\%$ human \texttt{ALWAYS\_PRESENT} overall). Human-Robust NSI correlates at $\rho\!=\!0.27$ on the range-restricted representative set and $\rho\!=\!0.69$ on the diagnostic set. A text-only audit rates the prompts natural, grounded and contrastive at $99.7$-$100\%$ and majority-flags one atom. Adding four anchor defects from a separate normalization pass, $5/300$ atoms need minor correction.

\noindent\textbf{Robustness:} Four analyses in three supplementary sections (\emph{Threshold Sensitivity and INVERSE Reachability}, \emph{Realization-Filter Robustness}, \emph{Seed-Variance Study}) test how far the findings depend on specific choices. \emph{(i) Thresholds:} over a $12$-point $(\theta_{\text{pos}},\theta_{\text{neg}})$ grid \texttt{INVERSE} never exceeds $0.93\%$, but class shares move substantially (\texttt{OTHER} spans $10$-$97\%$), so they are threshold-specific. The thresholds were set by hand, not calibrated on held-out data. Over $\xi\!\in\!\{0.2,0.3,0.4\}$, NSI shifts $\leq\!0.005$ and form $\times$ level rankings hold ($\tau\!\geq\!0.80$). \emph{(ii) Reachability:} swapping each \texttt{SUCCESS} atom's images relabels $100\%$ as \texttt{INVERSE}, so the region is reachable but narrow (only $\sim\!0.6\%$ of atoms have $C_j\!<\!0.3$). A threshold-free view finds $4.2$-$8.3\%$ of atoms with $\mathrm{NSI}_j\!<\!-0.10$. \emph{(iii) Filter:} retention is $\approx\!99.4\%$ overall and $\geq\!97.5\%$ per cell, and the form ordering and L2 trough survive unfiltered and common-support analyses. \emph{(iv) Seeds:} re-rendering $1{,}000$ atoms at $K\!=\!4$ leaves the aggregate unchanged (paired CI contains zero), with $93$-$94\%$ label stability and no \texttt{INVERSE}. Atom-bootstrap CIs separate the open ranking: FLUX.2 $[.0373,.0384]\!>\!$~Qwen-Image-2512 $[.0301,.0311]\!>\!$~SD~3.5 $[.0193,.0202]$.

% =============================================================================
\section{Conclusion}
\label{sec:conclusion}
We introduced T2I-NSI, the first benchmark-scale evaluation of negation in T2I generation. It factorizes $759{,}869$ paired prompts by linguistic form and target complexity, reduces probe-side confounds with canonical positive anchors, and decomposes failures with a six-way failure-mode taxonomy. The key finding is \emph{structural}: models almost never invert negation ($\texttt{INVERSE}\!=\!0.00\%$ over $450$K atom-model evaluations) but predominantly under-suppress it. A further large subset is too weakly rendered to label, and stronger closed models improve suppression without removing this signature. Human judgments, a matched-prompt comparison and threshold, filter and seed analyses support these conclusions, though exact class shares depend on the thresholds. Syntactic and relational negation are the hardest open-tier cases, and relational negation also scores lowest for GPT-Image-1.5 and Gemini-2.5-Flash-Image in exploratory results. Future work should target suppression directly, e.g.\ through paired-prompt training and scoped-negation paraphrasing.

% =============================================================================
% Limitations and Ethical Considerations
% =============================================================================
\section*{Limitations and Ethical Considerations}

\noindent\textbf{Probe-side bias and validity:} CLIP, InstructBLIP and GroundingDINO each inherit some negation blindness~\citep{alhamoud2025vision,zhang2025negvqa}. A probe that scores positive and negated images alike depresses NSI even when the generator did suppress the concept, which the Robust ensemble mitigates but does not eliminate. The probes are not uniformly conservative: humans read residual presence \emph{higher} than the probe ($0.78$ vs.\ $0.62$), so the metric can understate persistence while also understating success. We therefore report absolute scores as automated measurements. Inter-annotator agreement is limited (ordinal $\alpha\!=\!0.38$; one triad ICC$\,=\!-0.03$), as is five-way label agreement ($\kappa\!=\!0.09$). The low $\kappa$ comes mostly from one calibration offset ($283$ of the $359$ atoms humans call \texttt{ALWAYS\_PRESENT} fall in the metric's \texttt{OTHER}). The negative ICC comes from one triad that applied the scale inconsistently; the other two reach $0.79$ and $0.88$, and excluding it moves $\rho$ only from $0.27$ to $0.31$. Probe scales also differ. CLIP scores sit in a narrow band below $\theta_{\text{pos}}$ (10th-90th percentile $0.58$-$0.63$ on the $70\%$ of atoms with stored per-probe scores), so $C_j\!\geq\!\theta_{\text{pos}}$ effectively requires the other two probes to agree. VQA supplies most of the suppression signal: its mean positive-minus-negated difference is $0.09$-$0.16$, against $\leq\!0.004$ for CLIP and $\leq\!0.011$ in magnitude for the detector (on the $82\%$ of atoms with stored scores). Future versions should normalize probes before aggregation and calibrate thresholds on held-out human labels.

\noindent\textbf{Scope:} Open-weights evaluation covers three generators at full scale; the closed tier covers three at $N\!=\!228$-$1{,}200$, with uneven coverage (Midjourney~v7 is syntactic-only), so per-cell closed rankings are exploratory. The human study covers FLUX.2 and Qwen-Image-2512 only, not the closed-source models. Every prompt is LLM-generated, so T2I-NSI measures a constructed distribution rather than real user requests. Prompts are English-only and target visually grounded negation (non-visual predicates and relations not observable in one image are excluded by design), so scores do not generalize across languages or contexts. Although the main runs use $K\!=\!1$, the $K\!=\!4$ study confirms aggregates, orderings and taxonomy are seed-stable, though per-atom NSI retains SD $\approx\!0.05$.

\noindent\textbf{Data provenance and responsible use:} The catalog combines a curated noun list with the visually grounded subset of Synthetic Visual Genome, pruned by a triple-judge protocol that averages but does not eliminate shared model bias. \emph{Human subjects:} participation in the annotation study (nine annotators, three auditors) was voluntary and informed, no identifying information was collected, and only aggregate statistics are released. \emph{Data protection:} every prompt is model-generated from a curated object catalog containing no person names, and no identifying information was collected from annotators. We release the prompt corpus, the rendered images and the Basic and Robust scoring code on GitHub; exact regeneration is possible only for the locally run open-weights models. T2I-NSI is a diagnostic, not a safety guarantee: it certifies no single generation, and we release only visually grounded, non-harmful negations.

% =============================================================================
% Acknowledgments  (camera-ready only -- omit in any anonymous version)
% =============================================================================
\section*{Acknowledgments}
This research was supported by the IndiaAI Mission and Meta through the Srijan Centre of Excellence for Generative AI, and by the FAE team, AMD India, through GPU resources.
% =============================================================================
% Bibliography
% =============================================================================
{
    \small
    \bibliographystyle{ieeenat_fullname}
    \bibliography{main}

@String(CVPR   = {CVPR})

@String(ICCV   = {ICCV})

@String(ECCV   = {ECCV})

@String(WACV   = {WACV})

@String(NEURIPS = {NeurIPS})

@String(ICML   = {ICML})

@String(AAAI   = {AAAI})

@String(EMNLP  = {EMNLP})

@String(NAACL  = {NAACL})

@String(EACL   = {EACL})

@String(ACL    = {ACL})

@inproceedings{singh2025learning,
  title     = {Learning the Power of ``No'': Foundation Models with Negations},
  author    = {Singh, Jaisidh and Shrivastava, Ishaan and Vatsa, Mayank and Singh, Richa and Bharati, Aparna},
  booktitle = WACV,
  pages     = {7991--8001},
  year      = {2025}
}

@article{negation_taxonomy_2025,
  title   = {From No to Know: Taxonomy, Challenges, and Opportunities for Negation Understanding in Multimodal Foundation Models},
  author  = {Vatsa, Mayank and Bharati, Aparna and Mittal, Surbhi and Singh, Richa},
  journal = {arXiv preprint arXiv:2502.09645},
  year    = {2025}
}

@inproceedings{alhamoud2025vision,
  title     = {Vision-Language Models Do Not Understand Negation},
  author    = {Alhamoud, Kumail and Alshammari, Shaden and Tian, Yonglong and Li, Guohao and Torr, Philip H. S. and Kim, Yoon and Ghassemi, Marzyeh},
  booktitle = CVPR,
  pages     = {29612--29622},
  year      = {2025}
}

@inproceedings{ko2025bringing,
  title     = {Bringing {CLIP} to the Clinic: Dynamic Soft Labels and Negation-Aware Learning for Medical Analysis},
  author    = {Ko, Hanbin and Park, Chang-Min},
  booktitle = CVPR,
  pages     = {25897--25906},
  year      = {2025}
}

@inproceedings{zhang2025negvqa,
  title     = {{NegVQA}: Can Vision Language Models Understand Negation?},
  author    = {Zhang, Yuhui and Su, Yuchang and Liu, Yiming and Yeung-Levy, Serena},
  booktitle = {Findings of ACL},
  year      = {2025}
}

@inproceedings{so2025thunder,
  title     = {Thunder-{NUBench}: A Benchmark for {LLMs}' Sentence-Level Negation Understanding},
  author    = {So, Yeonkyoung and Lee, Gyuseong and Jung, Sungmok and Lee, Joonhak and Kang, JiA and Kim, Sangho and Lee, Jaejin},
  booktitle = {Findings of EACL},
  pages     = {4749--4793},
  year      = {2026}
}

@inproceedings{garcia-ferrero-etal-2023-dataset,
  title     = {This is not a Dataset: A Large Negation Benchmark to Challenge Large Language Models},
  author    = {Garc{\'\i}a-Ferrero, Iker and Altuna, Bego{\~n}a and Alvez, Javier and Gonzalez-Dios, Itziar and Rigau, German},
  booktitle = EMNLP,
  pages     = {8596--8615},
  year      = {2023}
}

@inproceedings{singh-etal-2023-nlms,
  title     = {{NLMs}: Augmenting Negation in Language Models},
  author    = {Singh, Rituraj and Kumar, Rahul and Sridhar, Vivek},
  booktitle = {Findings of EMNLP},
  pages     = {13104--13116},
  year      = {2023}
}

@inproceedings{rezaei2025making,
  title     = {Making Language Models Robust Against Negation},
  author    = {Rezaei, MohammadHossein and Blanco, Eduardo},
  booktitle = NAACL,
  year      = {2025}
}

@inproceedings{park2025know,
  title     = {Know ``No'' Better: A Data-Driven Approach for Enhancing Negation Awareness in {CLIP}},
  author    = {Park, Junsung and Lee, Jungbeom and Song, Jongyoon and Yu, Sangwon and Jung, Dahuin and Yoon, Sungroh},
  booktitle = ICCV,
  year      = {2025}
}

@inproceedings{bhatia2025learning,
  title     = {Answering Multimodal Exclusion Queries with Lightweight Sparse Disentangled Representations},
  author    = {{Prachi J} and Bhatia, Sumit and Bedathur, Srikanta},
  booktitle = {Int. Conf. on Innovative Concepts and Theories in Information Retrieval (ICTIR)},
  year      = {2025}
}

@inproceedings{radford2021clip,
  title     = {Learning Transferable Visual Models from Natural Language Supervision},
  author    = {Radford, Alec and Kim, Jong Wook and Hallacy, Chris and Ramesh, Aditya and Goh, Gabriel and Agarwal, Sandhini and Sastry, Girish and Askell, Amanda and Mishkin, Pamela and Clark, Jack and Krueger, Gretchen and Sutskever, Ilya},
  booktitle = ICML,
  pages     = {8748--8763},
  year      = {2021}
}

@inproceedings{dai2023instructblip,
  title={{InstructBLIP}: Towards General-purpose Vision-Language Models with Instruction Tuning},
  author={Dai, Wenliang and Li, Junnan and Li, Dongxu and Tiong, Anthony Meng Huat and Zhao, Junqi and Wang, Weisheng and Li, Boyang and Fung, Pascale and Hoi, Steven},
  booktitle = NEURIPS,
  year={2023}
}

@inproceedings{liu2024grounding,
  title     = {Grounding {DINO}: Marrying {DINO} with Grounded Pre-training for Open-Set Object Detection},
  author    = {Liu, Shilong and Zeng, Zhaoyang and Ren, Tianhe and Li, Feng and Zhang, Hao and Yang, Jie and Jiang, Qing and Li, Chunyuan and Yang, Jianwei and Su, Hang and Zhu, Jun and Zhang, Lei},
  booktitle = ECCV,
  year      = {2024}
}

@misc{flux2024,
  title        = {{FLUX.2-klein-9B} Model Card},
  author       = {{Black Forest Labs}},
  year         = {2026},
  howpublished = {\url{https://huggingface.co/black-forest-labs/FLUX.2-klein-9B}},
  note         = {Open-weights release; exact checkpoint evaluated in this work. Accessed 2026-08-29}
}

@inproceedings{esser2024scaling,
  title     = {Scaling Rectified Flow Transformers for High-Resolution Image Synthesis},
  author    = {Esser, Patrick and Kulal, Sumith and Blattmann, Andreas and Entezari, Rahim and M{\"u}ller, Jonas and Saini, Harry and Levi, Yam and Lorenz, Dominik and Sauer, Axel and Boesel, Frederic and others},
  booktitle = ICML,
  year      = {2024}
}

@misc{qwenimage2025,
  title        = {{Qwen-Image-2512} Model Card},
  author       = {{Alibaba Qwen Team}},
  year         = {2025},
  howpublished = {\url{https://huggingface.co/Qwen/Qwen-Image-2512}},
  note         = {Open-weights release; exact checkpoint evaluated in this work. Accessed 2026-08-29}
}

@misc{sd35turbo,
  title        = {{Stable Diffusion 3.5 Large Turbo} Model Card},
  author       = {{Stability AI}},
  year         = {2024},
  howpublished = {\url{https://huggingface.co/stabilityai/stable-diffusion-3.5-large-turbo}},
  note         = {Open-weights release; exact checkpoint evaluated in this work. Accessed 2026-08-29}
}

@misc{gptimage15,
  title        = {{GPT-Image-1.5}: Image Generation Model Documentation},
  author       = {{OpenAI}},
  year         = {2025},
  howpublished = {\url{https://platform.openai.com/docs/models/gpt-image-1.5}},
  note         = {Closed API; model identifier \texttt{gpt-image-1.5} as invoked in this work. Accessed 2026-08-29}
}

@misc{geminiflashimage,
  title        = {{Gemini 2.5 Flash Image}: Model Documentation},
  author       = {{Google DeepMind}},
  year         = {2025},
  howpublished = {\url{https://ai.google.dev/gemini-api/docs/models/gemini-2.5-flash-image}},
  note         = {Closed API; model identifier \texttt{gemini-2.5-flash-image} as invoked in this work. Accessed 2026-08-29}
}

@misc{midjourneyv7,
  title        = {{Midjourney} Version 7 Release Notes},
  author       = {{Midjourney}},
  year         = {2025},
  howpublished = {\url{https://docs.midjourney.com/}},
  note         = {Closed service; version evaluated in this work. Accessed 2026-08-29}
}

@misc{deepseekv32,
  title        = {{DeepSeek-V3.2} Model Card},
  author       = {{DeepSeek-AI}},
  year         = {2025},
  howpublished = {\url{https://huggingface.co/unsloth/DeepSeek-V3.2-GGUF/tree/main/UD-Q4_K_XL}},
  note         = {Open-weights release; served locally as the Unsloth \texttt{UD-Q4\_K\_XL} quantized GGUF build. Accessed 2026-08-29}
}

@misc{claudesonnet46,
  title        = {{Claude Sonnet 4.6} Model Documentation},
  author       = {{Anthropic}},
  year         = {2026},
  howpublished = {\url{https://docs.anthropic.com/en/docs/about-claude/models}},
  note         = {Closed API; used for closed-tier prompt generation. Accessed 2026-08-29}
}

@inproceedings{rombach2022high,
  title     = {High-Resolution Image Synthesis with Latent Diffusion Models},
  author    = {Rombach, Robin and Blattmann, Andreas and Lorenz, Dominik and Esser, Patrick and Ommer, Bj{\"o}rn},
  booktitle = CVPR,
  pages     = {10684--10695},
  year      = {2022}
}

@inproceedings{huang2023t2icompbench,
  title     = {{T2I-CompBench}: A Comprehensive Benchmark for Open-World Compositional Text-to-Image Generation},
  author    = {Huang, Kaiyi and Sun, Kaiyue and Xie, Enze and Li, Zhenguo and Liu, Xihui},
  booktitle = NEURIPS,
  year      = {2023}
}

@inproceedings{ghosh2023geneval,
  title     = {{GenEval}: An Object-Focused Framework for Evaluating Text-to-Image Alignment},
  author    = {Ghosh, Dhruba and Hajishirzi, Hannaneh and Schmidt, Ludwig},
  booktitle = NEURIPS,
  year      = {2023}
}

@inproceedings{cho2023dalleval,
  title     = {{DALL-Eval}: Probing the Reasoning Skills and Social Biases of Text-to-Image Generation Models},
  author    = {Cho, Jaemin and Zala, Abhay and Bansal, Mohit},
  booktitle = ICCV,
  year      = {2023}
}

@inproceedings{hu2023tifa,
  title     = {{TIFA}: Accurate and Interpretable Text-to-Image Faithfulness Evaluation with Question Answering},
  author    = {Hu, Yushi and Liu, Benlin and Kasai, Jungo and Wang, Yizhong and Ostendorf, Mari and Krishna, Ranjay and Smith, Noah A.},
  booktitle = ICCV,
  year      = {2023}
}

@inproceedings{hessel2021clipscore,
  title     = {{CLIPScore}: A Reference-Free Evaluation Metric for Image Captioning},
  author    = {Hessel, Jack and Holtzman, Ari and Forbes, Maxwell and Bras, Ronan Le and Choi, Yejin},
  booktitle = EMNLP,
  year      = {2021}
}

@inproceedings{singh2023divide,
  title     = {Divide, Evaluate, and Refine: Evaluating and Improving Text-to-Image Alignment with Iterative {VQA} Feedback},
  author    = {Singh, Jaskirat and Zheng, Liang},
  booktitle = NEURIPS,
  year      = {2023}
}

@article{cai2025tngclip,
  title   = {{TNG-CLIP}: Training-Time Negation Data Generation for Negation Awareness of {CLIP}},
  author  = {Cai, Yuliang and Thomason, Jesse and Rostami, Mohammad},
  journal = {arXiv preprint arXiv:2505.18434},
  year    = {2025}
}

@article{ranjbar2025spacevlm,
  title   = {{SpaceVLM}: Sub-Space Modeling of Negation in Vision-Language Models},
  author  = {Ranjbar, Sepehr Kazemi and Alhamoud, Kumail and Ghassemi, Marzyeh},
  journal = {arXiv preprint arXiv:2511.12331},
  year    = {2025}
}

@inproceedings{xiao2026clipglasses,
  title     = {Not Just What's There: Enabling {CLIP} to Comprehend Negated Visual Descriptions Without Fine-tuning},
  author    = {Xiao, Junhao and Wu, Zhiyu and Lin, Hao and Chen, Yi and Liu, Yahui and Zhao, Xiaoran and Wang, Zixu and He, Zejiang},
  booktitle = AAAI,
  year      = {2026}
}

@article{conwell2024relations,
  title   = {Relations, Negations, and Numbers: Looking for Logic in Generative Text-to-Image Models},
  author  = {Conwell, Colin and Tawiah-Quashie, Rupert and Ullman, Tomer},
  journal = {arXiv preprint arXiv:2411.17066},
  year    = {2024}
}

@inproceedings{sammani2026geometry,
  title     = {When Negation Is a Geometry Problem in Vision-Language Models},
  author    = {Sammani, Fawaz and Chamiti, Tzoulio and Gavrikov, Paul and Deligiannis, Nikos},
  booktitle = {CVPR Workshop on Multimodal Algorithmic Reasoning (MAR)},
  year      = {2026}
}

@inproceedings{hayes2025finegrain,
  title     = {{FineGRAIN}: Evaluating Failure Modes of Text-to-Image Models with Vision Language Model Judges},
  author    = {Hayes, Kevin David and Goldblum, Micah and Sehwag, Vikash and Somepalli, Gowthami and Panda, Ashwinee and Goldstein, Tom},
  booktitle = {NeurIPS Datasets and Benchmarks Track},
  year      = {2025}
}

@inproceedings{park2025synthetic,
  title     = {Synthetic Visual Genome},
  author    = {Park, Jae Sung and Ma, Zixian and Li, Linjie and Zheng, Chenhao and Hsieh, Cheng-Yu and Lu, Ximing and Chandu, Khyathi and Kong, Quan and Kobori, Norimasa and Farhadi, Ali and Choi, Yejin and Krishna, Ranjay},
  booktitle = CVPR,
  year      = {2025}
}
}

\end{document}